\documentclass[sigconf]{acmart}

\AtBeginDocument{%
  \providecommand\BibTeX{{%
    \normalfont B\kern-0.5em{\scshape i\kern-0.25em b}\kern-0.8em\TeX}}}

\setcopyright{acmlicensed}
\copyrightyear{2024}
\acmYear{2024}
\acmDOI{XXXXXXX.XXXXXXX}

\acmConference[MM'24]{Make sure to enter the correct
  conference title from your rights confirmation email}{October 28 - November 1,
  2024}{Melbourne, Australia.}
\acmISBN{978-1-4503-XXXX-X/18/06}

\begin{document}

\title{Deep Understanding of Soccer Match Videos}


\author{Shikun Xu, Yandong Zhu, Gen Li, Changhu Wang}
\email{}




\renewcommand{\shortauthors}{author name and author name, et al.}

\begin{abstract}
Soccer is one of the most popular sport worldwide, with live broadcasts frequently available for major matches. 
However, extracting detailed, frame-by-frame information on player actions from these videos remains a challenge.
Utilizing state-of-the-art computer vision technologies, our system can detect key objects such as soccer balls, players and referees. It also tracks the movements of players and the ball, recognizes player numbers, classifies scenes, and identifies highlights such as goal kicks. By analyzing live TV streams of soccer matches, our system can generate highlight GIFs, tactical illustrations, and diverse summary graphs of ongoing games.

Through these visual recognition techniques, we deliver a comprehensive understanding of soccer game videos, enriching the viewer's experience with detailed and insightful analysis.
\end{abstract}






\maketitle

\section{Introduction}

\begin{figure}[h]
\begin{center}
   \includegraphics[width=\linewidth]{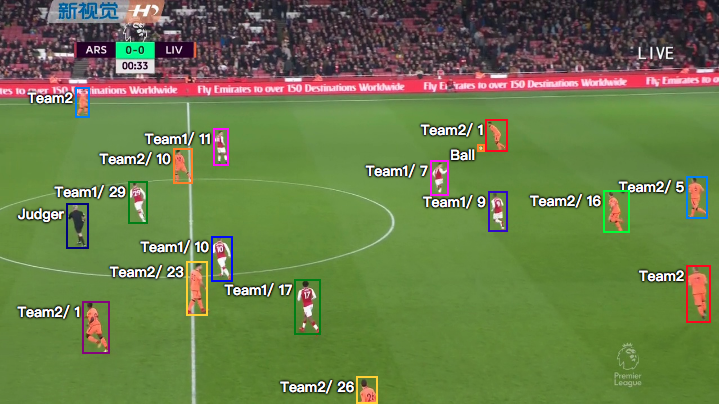}
\end{center}
   \caption{parsing results, including player detection and tracking, jersey number recognition, and team clustering. }
\label{fig:final}
\end{figure}

Video understanding encompasses a broad range of research fields, including video classification \cite{abu2016youtube}, video object detection\cite{lin2014microsoft}, video action recognition\cite{abu2016youtube}, video object segmentation\cite{perazzi2016benchmark} among others \cite{lin2014microsoft, zhang2015human}. Despite advances in modern computer vision techniques, achieving a deep understanding of live sports videos remains a formidable challenge due to the complexity of live matches. Factors such as small, low-resolution, and fast-moving players and ball, limited camera angles, and the diverse postures of players contribute to this difficulty. Additionally, current video processing algorithms primarily emphasize effectiveness, whereas efficiency is equally crucial in real-world applications, especially given the demands of live match analysis.

To automatically understand live soccer matches and generate meaningful summaries, we implement various vision algorithms tailored to soccer match videos. Firstly, we detect multiple objects on the field, including players, the ball, referees. In sports videos, objects are typically small, fast-moving, and of low resolution. Traditional image-based detectors \cite{ren2017faster, liu2016ssd, he2017mask} are either inefficient or perform poorly on small objects. We choose \cite{liu2016ssd} as our base model, modifying anchor settings and input shapes to improve small object detection.

To detect and track objects consecutively, we implemented a multi-object tracker primarily based on DeepSORT \cite{wojke2017simple}, equipped with a Kalman filter for motion estimation. Another key element in understanding soccer games is player identification. The classical solution of video person re-identification fails due to the similar appearances of players and a lack of training data. We partially solve this problem by recognizing jersey numbers. With player tracking results, we can identify most players in most frames.

Identifying a player's team and distinguishing players from referees is also necessary. We segment each person at pixel level and use color information to cluster individuals into two teams, referees, and goalkeepers. We also define, recognize, and localize various important actions of players and referees, such as shooting, sliding tackles, and yellow/red card warnings, which allows us to generate video highlights.

All the aforementioned information is based on 2D images from captured videos, lacking the positional context of the entire field, which is essential for high-level understanding of match processes. We define several semantic key points on the field and aim to detect them. Unlike other key point detection problems \cite{koestinger2011annotated, guo2016ms}, this task is more challenging due to large-scale variations and the invisibility of most points. We developed a two-stage CNN to localize key point positions and compute a homography transformation between the recorded field and a standard overhead view of the field.

Finally, our system generates highlight GIFs, tactical illustrations of highlights, and diverse summary graphs based on the aforementioned algorithms. Due to the careful design of our system, we can process live soccer match videos in real-time and deliver outputs to soccer fans.

To mitigate the influence of player and jersey appearance variations, we collected 164 soccer match videos from various leagues, including The Premier League and the AFC U-19 Championship. Most of these videos are recorded in at least 1080x720 resolution, with about 10\% in a higher resolution of 1920x1080. We manually labeled a portion of these videos to train and evaluate our algorithms.

\section{SYSTEM DESIGN}
In this section, we introduce the main functions of our system. The input to our system is a soccer match video file, and each function can generate results separately, which can be displayed as an option. Our system is designed to automatically process and understand live soccer match videos, comprising the following components:
\begin{itemize}
\item Object detection
\item Multi-object tracking
\item Jersey number recognition
\item Semantic segmentation
\item Team clustering
\item Highlight detection
\item Key point detection
\item Camera transformation
\end{itemize}

\begin{figure}[t]
\begin{center}
  \includegraphics[width=\linewidth]{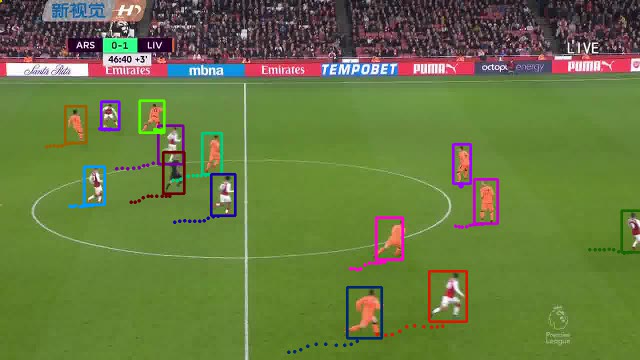}
\end{center}
  \caption{Illustration of the detection and tracking results.The bounding boxes represent detection results. The color tail represents the last 10 frames' positions of the player, which is calculated by the algorithm. }
\label{fig:track}
\end{figure}

\subsection{Object Detection and Tracking}
Considering both effectiveness and efficiency in real-world applications on soccer match videos, we use SSD \cite{liu2016ssd} instead of faster RCNN \cite{ren2017faster} or mask RCNN \cite{he2017mask}. Due to the relatively small sizes of the objects, we modify the default SSD configuration by setting the input shape to 640x640 and adjusting the anchor size range from 0.025 to 0.9. We fine-tune our model on open source dataset \cite{li2018jersey} with 215036 labeled bounding box, we use 10\% of the label as test set and achieving a mean Average Precision (mAP) of 78\%.

Our tracker is mainly based on DeepSORT \cite{wojke2017simple} which incorporates with a Kalman filter for motion estimation and an offline-trained appearance embedding network on person Re-ID dataset \cite{zheng2015scalable}. For the Kalman filter, we use the bounding boxes' coordinates and aspect ratio as inner states, which are updated when new measurements arrive. We associate inter-frame bounding boxes based on the cosine distance between their appearance features in a cascaded manner, as described in \cite{wojke2017simple}. Additionally, we use a hard metric that considers player information to reduce ID-switches, such as the team they belong to or their position in the formation. As shown in Figure~\ref{fig:track}, each person on the field, including referees and coaches, is detected and surrounded by a colorful bounding box. Each color represents an individual tracking result.

\subsection{Jersey Number Recognition}
We introduce a shallow CNN model with residual connection and integrate an STN\cite{jaderberg2015spatial} module on it. Based on the object detection results of all the players on the court, a VGG like CNN model is first used to classify these numbers on the deteced players’ images from 0 to 99. To localize the jersey number more precisely without involving another digit detector and extra consumption, we then improve the former network to an end-to-end framework by fusing with the spatial transformer network (STN) within convolution layer. To further improve the accuracy, we bring extra supervision to STN and upgrade the model to a semi-supervised multi-task learning system, by labeling a small portion of the number areas in the dataset by quadrangle. By integrating STN module and adding semi-supervised signal, we are able to boost the final performance from the CNN base model. 

Due to the variations of the players' poses and the angle of view of the camera, not all of the players' numbers can be captured in the frame. However, with our tracking results, we are able to inference the number of each player. The jersey number recognition results are shown in Figure~\ref{fig:final}. 

\begin{figure}[t]
\begin{center}
  \includegraphics[width=\linewidth]{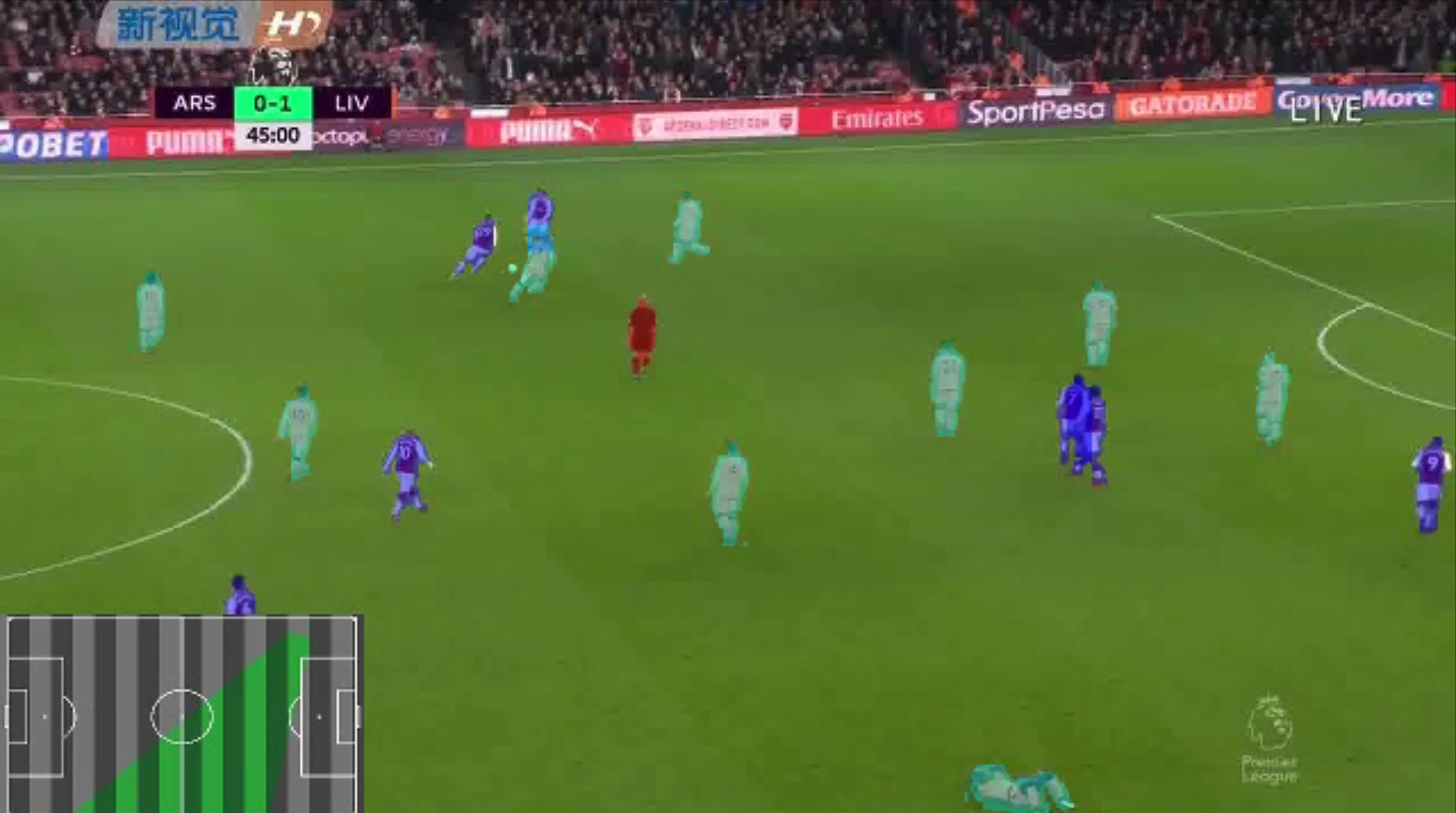}
\end{center}
  \caption{Illustration of the segmentation, team cluster and view transform results. The translucent mask represents each player's segmentation results and the color of mask represents different teams, including judges' color. The court map on the left bottom corner represents camera view field after transformed from the original image.}
\label{fig:cluster}
\end{figure}

\subsection{Segmentation and Team Clustering}
Soccer matches are often held on grass fields, which simplifies the task. We convert the image color space from RGB to HSV. By using an adaptive threshold, we can segment the players as precisely as a deep learning model but with much lower latency.

After segmentation, we compute the mean HSV and RGB values for each player's pixels. Since a player's jersey and shorts are usually different colors, we separately compute the mean values for each player's upper and lower parts. The team number is set to five to account for referees and goalkeepers from each team. We then use K-means to cluster all segmented players. The segmentation results are displayed with translucent colors, each representing the team to which the player belongs. The result is shown in Figure~\ref{fig:cluster}.

\subsection{Highlight Detection}
During a soccer match broadcast on TV, several camera shots may occur, including close shots, bird's-eye view shots, audience shots, and others. 

Highlight detection is approached as the classification of video clips into several categories, including shooting, corner kicks, penalties, free kicks, injuries, substitutions, and normal play. Each clip is eight seconds long, with one frame per second fed into a ResNet \cite{he2016deep} to extract deep features. These features are then aggregated using average pooling, and a softmax classifier is employed to produce the final classification results.

We manually label about 1000 samples for each class and achieving an accuracy of nearly 100\%.

\subsection{Key Point Detection}
Accurately determining the positions of the soccer ball and players is crucial for a high-level understanding of the game. To achieve this, we developed a two-stage CNN to localize key points and compute a homography transformation between the recorded field and a standard overhead view. We defined 17 key points on the field, including all the corners, the penalty area, and the center circle.

This task is particularly challenging due to significant scale variations and the invisibility of many key points. However, it is essential for accurately understanding player and ball positions during a match. Our method consists of two stages. In the first stage, the entire image is input into the CNN model to detect the key point positions and their visibility. In the second stage, we crop areas around the detected key points to refine their locations. The homography transformation is then applied to convert the 2D image coordinates to the corresponding field coordinates. As shown in Figure~\ref{fig:cluster}, this allows us to transform the entire field region in the camera view to field coordinates, providing accurate positional awareness on the court.

\begin{figure}[t]
\begin{center}
  \includegraphics[width=\linewidth]{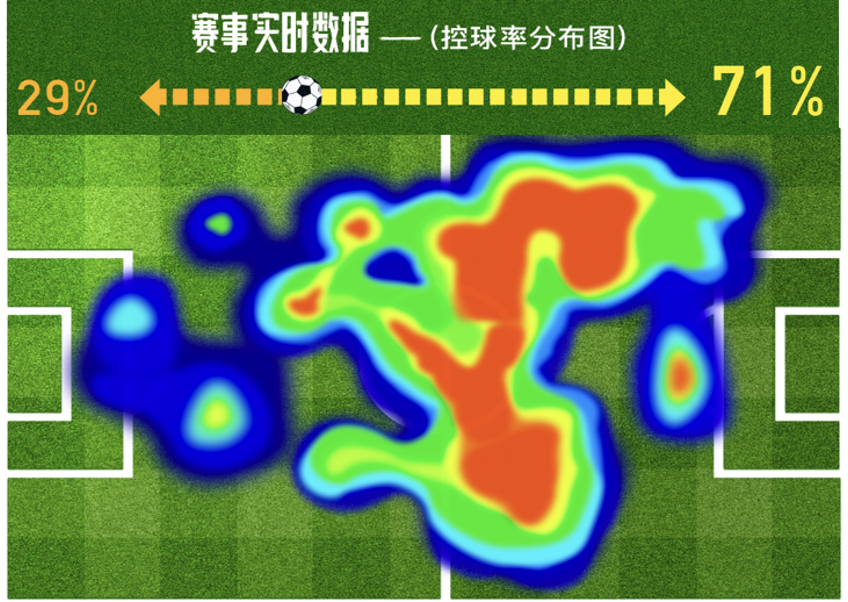}
\end{center}
  \caption{Illustration of the game summary, including the controlling rates, ball position heatmap, and team controlling distributes over the field.}
\label{fig:summary}
\end{figure}

\subsection{Game Summary}
By the techniques aforementioned, we could output several summary outputs, including the controlling rates, ball position heatmap, and team controlling distributes over the whole game. One summary output is illustrated in Figure~\ref{fig:summary}.

\section{CONCLUSIONS}
We implement several computer vision algorithms on soccer match videos in order to make computer understand it deeply. With our system, high level knowledge can be extracted from live match videos and diverse build-up computer vision techniques behind these knowledge are shown.










\bibliographystyle{ACM-Reference-Format}
\bibliography{sample-base}


\begin{thebibliography}{14}


\ifx \showCODEN    \undefined \def \showCODEN     #1{\unskip}     \fi
\ifx \showDOI      \undefined \def \showDOI       #1{#1}\fi
\ifx \showISBNx    \undefined \def \showISBNx     #1{\unskip}     \fi
\ifx \showISBNxiii \undefined \def \showISBNxiii  #1{\unskip}     \fi
\ifx \showISSN     \undefined \def \showISSN      #1{\unskip}     \fi
\ifx \showLCCN     \undefined \def \showLCCN      #1{\unskip}     \fi
\ifx \shownote     \undefined \def \shownote      #1{#1}          \fi
\ifx \showarticletitle \undefined \def \showarticletitle #1{#1}   \fi
\ifx \showURL      \undefined \def \showURL       {\relax}        \fi
\providecommand\bibfield[2]{#2}
\providecommand\bibinfo[2]{#2}
\providecommand\natexlab[1]{#1}
\providecommand\showeprint[2][]{arXiv:#2}

\bibitem[Abu-El-Haija et~al\mbox{.}(2016)]%
        {abu2016youtube}
\bibfield{author}{\bibinfo{person}{Sami Abu-El-Haija}, \bibinfo{person}{Nisarg Kothari}, \bibinfo{person}{Joonseok Lee}, \bibinfo{person}{Paul Natsev}, \bibinfo{person}{George Toderici}, \bibinfo{person}{Balakrishnan Varadarajan}, {and} \bibinfo{person}{Sudheendra Vijayanarasimhan}.} \bibinfo{year}{2016}\natexlab{}.
\newblock \showarticletitle{Youtube-8m: A large-scale video classification benchmark}.
\newblock \bibinfo{journal}{\emph{arXiv preprint arXiv:1609.08675}} (\bibinfo{year}{2016}).
\newblock


\bibitem[Guo et~al\mbox{.}(2016)]%
        {guo2016ms}
\bibfield{author}{\bibinfo{person}{Yandong Guo}, \bibinfo{person}{Lei Zhang}, \bibinfo{person}{Yuxiao Hu}, \bibinfo{person}{Xiaodong He}, {and} \bibinfo{person}{Jianfeng Gao}.} \bibinfo{year}{2016}\natexlab{}.
\newblock \bibinfo{booktitle}{\emph{Ms-celeb-1m: A dataset and benchmark for large-scale face recognition}}.
\newblock Springer.
\newblock


\bibitem[He et~al\mbox{.}(2017)]%
        {he2017mask}
\bibfield{author}{\bibinfo{person}{Kaiming He}, \bibinfo{person}{Georgia Gkioxari}, \bibinfo{person}{Piotr Doll{\'a}r}, {and} \bibinfo{person}{Ross Girshick}.} \bibinfo{year}{2017}\natexlab{}.
\newblock \bibinfo{booktitle}{\emph{Mask r-cnn}}.
\newblock IEEE.
\newblock


\bibitem[He et~al\mbox{.}(2016)]%
        {he2016deep}
\bibfield{author}{\bibinfo{person}{Kaiming He}, \bibinfo{person}{Xiangyu Zhang}, \bibinfo{person}{Shaoqing Ren}, {and} \bibinfo{person}{Jian Sun}.} \bibinfo{year}{2016}\natexlab{}.
\newblock \bibinfo{booktitle}{\emph{Deep residual learning for image recognition}}.
\newblock


\bibitem[Jaderberg et~al\mbox{.}(2015)]%
        {jaderberg2015spatial}
\bibfield{author}{\bibinfo{person}{Max Jaderberg}, \bibinfo{person}{Karen Simonyan}, \bibinfo{person}{Andrew Zisserman}, {et~al\mbox{.}}} \bibinfo{year}{2015}\natexlab{}.
\newblock \bibinfo{booktitle}{\emph{Spatial transformer networks}}.
\newblock


\bibitem[Koestinger et~al\mbox{.}(2011)]%
        {koestinger2011annotated}
\bibfield{author}{\bibinfo{person}{Martin Koestinger}, \bibinfo{person}{Paul Wohlhart}, \bibinfo{person}{Peter~M Roth}, {and} \bibinfo{person}{Horst Bischof}.} \bibinfo{year}{2011}\natexlab{}.
\newblock \bibinfo{booktitle}{\emph{Annotated facial landmarks in the wild: A large-scale, real-world database for facial landmark localization}}.
\newblock IEEE.
\newblock


\bibitem[Li et~al\mbox{.}(2018)]%
        {li2018jersey}
\bibfield{author}{\bibinfo{person}{Gen Li}, \bibinfo{person}{Shikun Xu}, \bibinfo{person}{Xiang Liu}, \bibinfo{person}{Lei Li}, {and} \bibinfo{person}{Changhu Wang}.} \bibinfo{year}{2018}\natexlab{}.
\newblock \showarticletitle{Jersey number recognition with semi-supervised spatial transformer network}. In \bibinfo{booktitle}{\emph{Proceedings of the IEEE conference on computer vision and pattern recognition workshops}}. \bibinfo{pages}{1783--1790}.
\newblock


\bibitem[Lin et~al\mbox{.}(2014)]%
        {lin2014microsoft}
\bibfield{author}{\bibinfo{person}{Tsung-Yi Lin}, \bibinfo{person}{Michael Maire}, \bibinfo{person}{Serge Belongie}, \bibinfo{person}{James Hays}, \bibinfo{person}{Pietro Perona}, \bibinfo{person}{Deva Ramanan}, \bibinfo{person}{Piotr Doll{\'a}r}, {and} \bibinfo{person}{C~Lawrence Zitnick}.} \bibinfo{year}{2014}\natexlab{}.
\newblock \bibinfo{booktitle}{\emph{Microsoft coco: Common objects in context}}.
\newblock Springer.
\newblock


\bibitem[Liu et~al\mbox{.}(2016)]%
        {liu2016ssd}
\bibfield{author}{\bibinfo{person}{Wei Liu}, \bibinfo{person}{Dragomir Anguelov}, \bibinfo{person}{Dumitru Erhan}, \bibinfo{person}{Christian Szegedy}, \bibinfo{person}{Scott Reed}, \bibinfo{person}{Cheng-Yang Fu}, {and} \bibinfo{person}{Alexander~C Berg}.} \bibinfo{year}{2016}\natexlab{}.
\newblock \bibinfo{booktitle}{\emph{Ssd: Single shot multibox detector}}.
\newblock


\bibitem[Perazzi et~al\mbox{.}(2016)]%
        {perazzi2016benchmark}
\bibfield{author}{\bibinfo{person}{Federico Perazzi}, \bibinfo{person}{Jordi Pont-Tuset}, \bibinfo{person}{Brian McWilliams}, \bibinfo{person}{Luc Van~Gool}, \bibinfo{person}{Markus Gross}, {and} \bibinfo{person}{Alexander Sorkine-Hornung}.} \bibinfo{year}{2016}\natexlab{}.
\newblock \showarticletitle{A benchmark dataset and evaluation methodology for video object segmentation}. In \bibinfo{booktitle}{\emph{Proceedings of the IEEE Conference on Computer Vision and Pattern Recognition}}. \bibinfo{pages}{724--732}.
\newblock


\bibitem[Ren et~al\mbox{.}(2017)]%
        {ren2017faster}
\bibfield{author}{\bibinfo{person}{Shaoqing Ren}, \bibinfo{person}{Kaiming He}, \bibinfo{person}{Ross Girshick}, {and} \bibinfo{person}{Jian Sun}.} \bibinfo{year}{2017}\natexlab{}.
\newblock \bibinfo{booktitle}{\emph{Faster R-CNN: towards real-time object detection with region proposal networks}}.
\newblock


\bibitem[Wojke et~al\mbox{.}(2017)]%
        {wojke2017simple}
\bibfield{author}{\bibinfo{person}{Nicolai Wojke}, \bibinfo{person}{Alex Bewley}, {and} \bibinfo{person}{Dietrich Paulus}.} \bibinfo{year}{2017}\natexlab{}.
\newblock \bibinfo{booktitle}{\emph{Simple online and realtime tracking with a deep association metric}}.
\newblock


\bibitem[Zhang and Shah(2015)]%
        {zhang2015human}
\bibfield{author}{\bibinfo{person}{Dong Zhang} {and} \bibinfo{person}{Mubarak Shah}.} \bibinfo{year}{2015}\natexlab{}.
\newblock \showarticletitle{Human pose estimation in videos}. In \bibinfo{booktitle}{\emph{Proceedings of the IEEE International Conference on Computer Vision}}. \bibinfo{pages}{2012--2020}.
\newblock


\bibitem[Zheng et~al\mbox{.}(2015)]%
        {zheng2015scalable}
\bibfield{author}{\bibinfo{person}{Liang Zheng}, \bibinfo{person}{Liyue Shen}, \bibinfo{person}{Lu Tian}, \bibinfo{person}{Shengjin Wang}, \bibinfo{person}{Jingdong Wang}, {and} \bibinfo{person}{Qi Tian}.} \bibinfo{year}{2015}\natexlab{}.
\newblock \showarticletitle{Scalable person re-identification: A benchmark}. In \bibinfo{booktitle}{\emph{Proceedings of the IEEE International Conference on Computer Vision}}. \bibinfo{pages}{1116--1124}.
\newblock


\end{thebibliography}

\end{document}